\documentclass[11pt]{article}
\usepackage{coling2014}
\usepackage{times}
\usepackage{url}
\usepackage{latexsym}
\usepackage{graphicx}
\usepackage{mdwlist}

\title{Clinical TempEval}

\author{Steven Bethard \\
  University of Alabama at Birmingham \\
  Birmingham, AL 35294 \\
  USA \\
  {\tt bethard@cis.uab.edu} \\\And
  Leon Derczynski \\
  University of Sheffield \\
  Sheffield, S1 4DP \\
  UK \\
  {\tt leon@dcs.shef.ac.uk} \\\AND
  James Pustejovsky \\
  Brandeis University \\
  Waltham, MA 02453 \\
  USA \\
  {\tt jamesp@cs.brandeis.edu} \\\And
  Marc Verhagen \\
  Brandeis University \\
  Waltham, MA 02453 \\
  USA \\
  {\tt marc@cs.brandeis.edu} \\}

\date{}

\begin{document}
\maketitle
\begin{abstract}
We describe the Clinical TempEval task which is currently in preparation for the SemEval-2015 evaluation exercise.
This task involves identifying and describing events, times and the relations between them in clinical text.
Six discrete subtasks are included, focusing on recognising mentions of times and events, describing those mentions for both entity types, identifying the relation between an event and the document creation time, and identifying narrative container relations.
\end{abstract}

\section{Summary}
ClinicalTempEval will bring the temporal information extraction tasks of previous TempEvals to the clinical domain, using clinical notes of colon cancer patients from the Mayo Clinic. Our definitions of events and times are based on those in TimeML~\cite{pustejovsky2003timeml,pustejovsky2010iso}. ClinicalTempEval will provide the following temporal annotation sub-tasks:

\begin{itemize*}
\item TS: identifying the spans of time expressions
\item ES: Identifying the spans of event expressions
\item TA: identifying the attributes of time expressions (type = DATE / TIME / DURATION / QUANTIFIER / PREPOSTEXP / SET; value = TIMEX3.val) -- see~\cite{ferro2005tides} for additional details of the TIMEX specification
\item EA: identifying the attributes of event expressions (type = NA/ASPECTUAL/EVIDENTIAL; polarity = POS / NEG; degree = NA / MOST / LITTLE; modality = ACTUAL / HEDGED / HYPOTHETICAL / GENERIC)
\item DR: identifying the relation between an event and the document creation time (docTimeRel = BEFORE / OVERLAP / AFTER / BEFORE-OR-OVERLAP)
\item CR: identifying narrative container relations (CONTAINS a.k.a. INCLUDES)
\end{itemize*}

These sub-tasks are largely common to previous similar exercises. They will be presented in three scenarios, detailed in Section~\ref{sec:eval}.

\section{Motivation}
The TempEval task has, since 2007, provided a focus for research on temporal information extraction~\cite{verhagen2009tempeval,verhagen2010semeval,uzzaman2013semeval}. The automatic identiﬁcation of all temporal referring expressions, events and temporal relations within a text is the ultimate aim of research in this area. As a result of previous TempEvals, we have discovered much about temporal information extraction, identifying the difficulties in the area and contributing new resources.

TempEval’s information extraction exercises have been presented as discrete, well-defined tasks, with automatic and quantitative evaluation of approaches a key part. We continue this format, looking at the most difficult parts of temporal IE in a formal manner. This makes evaluation rapid, reliable and repeatable, focused on the key parts of the problem instead of potentially harder-to-assess situations like, for example, free-form question answering or event-based summarisation.

ClinicalTempEval extends TempEval to the clinical domain. This move is for two key reasons. First, concentrating on newswire constrains our understanding of time in language to a particular range of expressions, event types and timeframes. Second, there is great interest in temporal information extraction in this domain, and great utility to be had in solving it. For example, in 2012 a traditional clinical NLP challenge (i2b2) ran their shared task on just this problem~\cite{sun2013evaluating}. We have already acquired annotations over sharable clinical texts for the task.

\section{Data}
The THYME project\footnote{See \url{http://projectreporter.nih.gov/project_info_description.cfm?aid=8138604\&icde=10245671} } is currently annotating times, events and temporal relations in clinical notes following guidelines derived from ISO TimeML for the THYME project. (The i2b2 guidelines were derived from the THYME guidelines and are essentially a subset of the THYME guidelines.) The annotation pipeline is as follows:

\begin{enumerate*}
\item Annotators identify time and event expressions, along with their attributes (for events, attributes include the temporal relation to the document creation time)
\item Adjudicators revise and finalize the time and event expressions and their attributes
\item Annotators identify temporal relations between pairs of events and between events and times
\item Adjudicators revise and finalize the temporal relations
\end{enumerate*}

Currently, 232 notes from 87 patients have been annotated, with over 30000 events, 2500 times and 9000 narrative container relations~\cite{pustejovsky2011increasing,miller2013discovering}. We anticipate around 600 notes from 200 patients by the time the TempEval 2015 training data is released.

The \textsc{Pheme} project\footnote{See \url{http://www.pheme.eu}} which starts January 2014 involves annotation of spatial and temporal aspects of non-newswire text (with a focus on social media). This project will provide annotations for time expression values.

For Clinical TempEval, we will use splits at patient record level. This means that patient data will not leak across datasets. One half of the patient records will be used as training data, one quarter as development data, and the final quarter as the test set. This gives a test set roughly the size of half of the TimeBank corpus~\cite{pustejovsky2003timebank}.

\section{Data Use Agreements}
All clinical notes have been de-identified, but access to the TempEval 2015 data will still require participants to sign a data use agreement with the Mayo Clinic, to ensure that the data is handled appropriately. After the competition, the data set will be available to other researchers (though again, requiring a data use agreement).

\section{Evaluation}
\label{sec:eval}
We envision three different evaluation setups:
\begin{enumerate*}
\item Only the plain text is given
\item Manually annotated event and time expression spans are given
\item Manually annotated event and time expression spans and attributes are given
\end{enumerate*}

Evaluation for each setup will be as follows:
\begin{enumerate*}
\item Only the plain text is given
\begin{itemize*}
\item TS, ES: precision, recall and F1
\item TA, EA: precision, recall and F1 for each attribute, and an overall precision, recall and F1 where a time/event is marked correct only if all attributes are correct
\item DR: precision, recall and F1
\item CR: precision, recall and F1, and closure-based precision, recall and F1, where temporal closure is run to infer additional relations on both the system and the reference relations and scores are calculated on the post-closure relations.
\end{itemize*}
\item Manually annotated event and time expression spans are given
\begin{itemize*}
\item TA, EA: accuracy for each attribute, and an overall accuracy where a time/event is marked correct only if all attributes are correct
\item DR: accuracy
\item CR: precision, recall and F1, and closure-based precision, recall and F1.
\end{itemize*}
\item Manually annotated event and time expression spans and attributes are given
\begin{itemize*}
\item DR: accuracy
\item CR: precision, recall and F1, and closure-based precision, recall and F1.
\end{itemize*}
\end{enumerate*}

\section{Resources Required}
Annotation costs are covered by the THYME project. There is already sufficient data available now to run a shared task, but annotation is ongoing, and we will make available whatever has been fully annotated and adjudicated at the time of the TempEval 2015 training data release.

The main resource that still needs to be developed is the evaluation scripts, used by the official evaluation, and also in a form that can be distributed to participants.

\section{Baseline Systems}
We will provide several baseline systems to compare against, such as:

\begin{itemize*}
\item For TS, ES, TA and EA: If an event/time was seen in the training data and it’s seen in the test data, mark it as an event/time and give identical attributes to whatever it had in the training data
\item For DR: the most common class and/or a memorization baseline like above
\item For CR: link each event to the closest time expression in the same sentence
\end{itemize*}

\section{Organizers}
\begin{itemize*}
\item Steven Bethard $<$bethard@cis.uab.edu$>$, The University of Alabama at Birmingham
\item Leon Derczynski $<$leon@dcs.shef.ac.uk$>$, The University of Sheffield
\item James Pustejovsky $<$jamesp@cs.brandeis.edu$>$, Brandeis University
\item Marc Verhagen $<$marc@cs.brandeis.edu$>$, Brandeis University
\end{itemize*}

\section*{Acknowledgments}

Clinical TempEval is supported by Temporal History of Your Medical Events (THYME), NRL R01LM010090, and EU FP7 grant agreement No. 611233, Pheme.

\bibliographystyle{aclabbrv}
\bibliography{clinte-proposal}

\begin{thebibliography}{}

\bibitem[\protect\citename{Ferro \bgroup et al.\egroup }2005]{ferro2005tides}
L.~Ferro, L.~Gerber, I.~Mani, B.~Sundheim, and G.~Wilson.
\newblock 2005.
\newblock Tides 2005 standard for the annotation of temporal expressions.

\bibitem[\protect\citename{Miller \bgroup et al.\egroup
  }2013]{miller2013discovering}
T.~A. Miller, S.~Bethard, D.~Dligach, S.~Pradhan, C.~Lin, and G.~K. Savova.
\newblock 2013.
\newblock Discovering narrative containers in clinical text.
\newblock {\em ACL 2013}, page~18.

\bibitem[\protect\citename{Pustejovsky and
  Stubbs}2011]{pustejovsky2011increasing}
J.~Pustejovsky and A.~Stubbs.
\newblock 2011.
\newblock Increasing informativeness in temporal annotation.
\newblock In {\em Proceedings of the 5th Linguistic Annotation Workshop}, pages
  152--160. Association for Computational Linguistics.

\bibitem[\protect\citename{Pustejovsky \bgroup et al.\egroup
  }2003a]{pustejovsky2003timeml}
J.~Pustejovsky, J.~M. Castano, R.~Ingria, R.~Sauri, R.~J. Gaizauskas,
  A.~Setzer, G.~Katz, and D.~R. Radev.
\newblock 2003a.
\newblock Timeml: Robust specification of event and temporal expressions in
  text.
\newblock {\em New directions in question answering}, 3:28--34.

\bibitem[\protect\citename{Pustejovsky \bgroup et al.\egroup
  }2003b]{pustejovsky2003timebank}
J.~Pustejovsky, P.~Hanks, R.~Sauri, A.~See, R.~Gaizauskas, A.~Setzer, D.~Radev,
  B.~Sundheim, D.~Day, L.~Ferro, et~al.
\newblock 2003b.
\newblock The {TimeBank} corpus.
\newblock In {\em Corpus linguistics}, volume 2003, page~40.

\bibitem[\protect\citename{Pustejovsky \bgroup et al.\egroup
  }2010]{pustejovsky2010iso}
J.~Pustejovsky, K.~Lee, H.~Bunt, and L.~Romary.
\newblock 2010.
\newblock Iso-timeml: An international standard for semantic annotation.
\newblock In {\em LREC}.

\bibitem[\protect\citename{Sun \bgroup et al.\egroup }2013]{sun2013evaluating}
W.~Sun, A.~Rumshisky, and O.~Uzuner.
\newblock 2013.
\newblock Evaluating temporal relations in clinical text: 2012 i2b2 challenge.
\newblock {\em Journal of the American Medical Informatics Association},
  20(5):806--813.

\bibitem[\protect\citename{UzZaman \bgroup et al.\egroup
  }2013]{uzzaman2013semeval}
N.~UzZaman, H.~Llorens, L.~Derczynski, M.~Verhagen, J.~Allen, and
  J.~Pustejovsky.
\newblock 2013.
\newblock Semeval-2013 task 1: Tempeval-3: Evaluating time expressions, events,
  and temporal relations.
\newblock In {\em Second joint conference on lexical and computational
  semantics (* SEM)}, volume~2, pages 1--9.

\bibitem[\protect\citename{Verhagen \bgroup et al.\egroup
  }2009]{verhagen2009tempeval}
M.~Verhagen, R.~Gaizauskas, F.~Schilder, M.~Hepple, J.~Moszkowicz, and
  J.~Pustejovsky.
\newblock 2009.
\newblock The tempeval challenge: identifying temporal relations in text.
\newblock {\em Language Resources and Evaluation}, 43(2):161--179.

\bibitem[\protect\citename{Verhagen \bgroup et al.\egroup
  }2010]{verhagen2010semeval}
M.~Verhagen, R.~Sauri, T.~Caselli, and J.~Pustejovsky.
\newblock 2010.
\newblock Semeval-2010 task 13: Tempeval-2.
\newblock In {\em Proc. SemEval}, pages 57--62.

\end{thebibliography}

\end{document}